\documentclass[10pt,twocolumn,letterpaper]{article}

\usepackage{iccv}
\usepackage{times}
\usepackage{epsfig}
\usepackage{graphicx}
\usepackage{amsmath}
\usepackage{amssymb}
\usepackage{multirow}

\usepackage{array}
\newcolumntype{C}[1]{>{\centering\let\newline\\\arraybackslash\hspace{0pt}}m{#1}}

\usepackage[breaklinks=true,bookmarks=false]{hyperref}

\iccvfinalcopy 

\def\iccvPaperID{****} 

\ificcvfinal\fi

\newcommand{\parsection}[1]{\noindent\textbf{#1}}

\begin{document}
	
	\title{AIM 2019 Challenge on Real-World Image Super-Resolution: \\Methods and Results}
	
	\author{Andreas Lugmayr \and Martin Danelljan \and Radu Timofte \and Manuel Fritsche \and Shuhang Gu\and Kuldeep Purohit \and Praveen Kandula \and Maitreya Suin \and A N Rajagopalan \and Nam Hyung Joon \and Yu Seung Won \and Guisik Kim \and Dokyeong Kwon \and Chih-Chung Hsu \and Chia-Hsiang Lin  \and Yuanfei Huang \and Xiaopeng Sun \and Wen Lu \and Jie Li \and Xinbo Gao \and Sefi Bell-Kligler}
	
	\maketitle
	\ificcvfinal\thispagestyle{empty}\fi

	\begin{abstract}
		This paper reviews the AIM 2019 challenge on real world super-resolution. It focuses on the participating methods and final results. The challenge addresses the real world setting, where paired true high and low-resolution images are unavailable. For training, only one set of source input images is therefore provided in the challenge. In \emph{Track~1: Source Domain} the aim is to super-resolve such images while preserving the low level image characteristics of the source input domain. In \emph{Track~2: Target Domain} a set of high-quality images is also provided for training, that defines the output domain and desired quality of the super-resolved images. To allow for quantitative evaluation, the source input images in both tracks are constructed using artificial, but realistic, image degradations. The challenge is the first of its kind, aiming to advance the state-of-the-art and provide a standard benchmark for this newly emerging task. In total 7 teams competed in the final testing phase, demonstrating new and innovative solutions to the problem.
	\end{abstract}
\let\thefootnote\relax\footnotetext{
A. Lugmayr (andreas.lugmayr@vision.ee.ethz.ch, ETH Zurich), M. Danelljan, and R. Timofte are the AIM 2019 challenge organizers, while the other authors participated in the challenge.\\
Appendix A contains the authors’ teams and affiliations.\\
AIM webpage: \url{http://www.vision.ee.ethz.ch/aim19/}
}
	
\begin{figure}[t]
\centering%
	\newcommand{\wid}{.5\linewidth}
	\includegraphics[width=\wid]{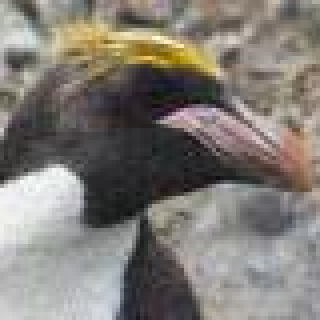}%
	\includegraphics[width=\wid]{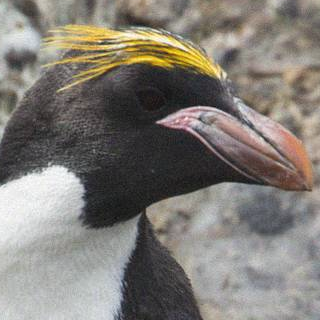}\vspace{-1mm}
	\resizebox{\linewidth}{!}{
		\begin{tabular}{C{4cm} C{4cm}}
			Input LR Track 1 &  Ground Truth HR Track 1
		\end{tabular}
	}\\
	\vspace{1mm}
	\includegraphics[width=\wid]{figures/Validation/12x_crop_NN4.png}%
	\includegraphics[width=\wid]{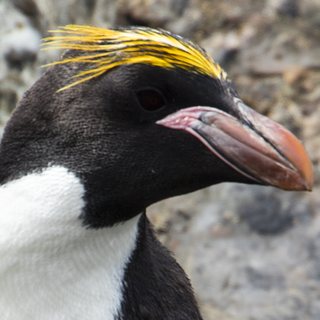}\vspace{-1mm}
	\resizebox{\linewidth}{!}{
		\begin{tabular}{C{4cm} C{4cm}}
			Input LR Track 2 &  Ground Truth HR Track 2
		\end{tabular}
	}
	\caption{Visual example of the input LR images and ground truth HR images used in the challenge. In both tracks, the input generated with a degradation operation that is unknown to the participants to simulate the real-world SR case where HR ground truth is unavailable. In Track 1: Source Domain, the aim is to generate an HR image with preserved low-level image characteristics. In Track 2: Target Domain, the goal is to achieve a clean image.}
	\vspace{-4mm}
	\label{fig:intro}
\end{figure}

\section{Introduction}

Single image Super-Resolution (SR) is the task of increasing the resolution of a given image. It has been a popular research topic for decades \cite{irani1991improving, freeman2002example, park2003super, yang2010image, huang2015single, dong2014learning, dong2016image, kim2016accurate, lai2017deep, lim2017EDSR, fan2017balanced, ahn2018fast, ahn2018image, haris2018deep, huang2018densely,AIM2019ESR} due to its many applications. It is well known that the SR task is a highly ill-posed inverse problem due to the incomplete information available in the given low-resolution image, which generates to a high-dimensional solution space of corresponding SR images. It is therefore essential to construct or learn priors to guide the super-resolution process itself.
Resent works have successfully tackled this problem by deploying deep Convolutional Neural Networks (CNNs). These have shown capable of learning powerful priors of image content and low-level characteristics. In particular, when combined with adversarial training \cite{ledig2017photo, wang2018esrgan}, SR networks can produce accurate and natural looking image details, although not without failure cases.

While deep learning based SR methods achieve current state-of-the-art, they need of large quantities of training data. Most current approaches rely on low and high-resolution image pairs to train the network in a fully supervised manner. However, such image pairs are not available in real-world applications, where images originate from a particular camera sensor. To circumvent this fact, the conventional approach has been to apply simple bicubic downsampling to artificially generate corresponding LR images. This strategy unfortunately introduces significant artifacts, by severely reducing the sensor noise and affecting other natural image characteristics. Super-resolution networks trained on such bicubic images therefore often struggle to generalize to natural images. This poses a fundamental challenge, calling for alternative SR approaches that can be learned without paired LR-HR supervision.

The blind super-resolution setting \cite{michaeli2013nonparametric, gu2019blind,begin2004blind} only partially addresses the aforementioned problem by assuming an unknown down-sampling kernel, but it still relies on paired examples for training.
Very recent works \cite{zhang2019zoom, chen2019camera, Cai_2019_CVPR_Workshops} propose strategies to capture real LR-HR image pairs. While these methods are helpful, they rely on complicated data collection procedures, requiring specialized hardware, that is difficult and expensive to scale. Moreover, it cannot be applied to old photo content. This challenge therefore focuses on the fully unsupervised super-resolution case, similar to the setting employed in many recent works \cite{yuan2018unsupervised,kim2018task,bulat2018learn,lugmayrICCVW2019}, Where no reference high-resolution images are available for training.

The AIM 2019 Challenge on Real-World Image Super-Resolution aims to stimulate research in the aforementioned direction by introducing a new benchmark dataset and protocol. In contrast to the conventional and blind SR setting, no paired LR-HR images are provided during training. To allow quantitative evaluation, we apply artificial but
realistic image degradations that are unknown (see figure~\ref{fig:intro} for examples). The dataset is constructed based on the popular DIV2K \cite{timofte2017ntire} and Flickr2K~\cite{wang2018esrgan} images. We evaluate and analyze the participating approaches based on classcal image quality metrics PSNR and SSIM, as well as the learned LPIPS \cite{zhang2018unreasonable} metric. Moreover, we base the final ranking on a human perceptual study.

	\section{AIM 2019 Challenge}

The goals of the AIM 2019 Challenge on Real-World Image Super-Resolution is to (i) promote research into weak and unsupervised learning approached for SR, that are applicable to the real-world settings; (ii) provide a common benchmark protocol and dataset; and (iii) probe the current state-of-the-art in the field.

\subsection{RealWorld SR Dataset}

We adopt the dataset and benchmark protocol recently introduced by Lugmayr et al.~\cite{lugmayrICCVW2019}, that allows for quantitative benchmarking of real-world super-resolution approaches.

\parsection{Degradation}
We simulate the real-world scenario by applying synthetic but realistic degradations to clean high-quality images. This models any sort of noise or corruption that might affect the image during the acquisition. However, not that the purpose is not to pursue the most realistic degradation, but to force the participants to employ the source domain data for learning. The degradation operation is unknown to the participants in the challenge. According to the rules of the challenge, the participants were not permitted to try to reverse-engineer or with hand-crafted algorithms construct similar-looking degradation artifacts. It was however allowed to try to \emph{learn} the degradation operator using generic techniques (such as deep networks) that can be applied to any other sort of degradations or source of natural images.

\parsection{Data}
We construct a dataset of source (\ie input) domain training images $\mathcal{X}_\text{train} = \{x_i\}$ by directly applying the degradation operation to the Flickr2K~\cite{wang2018esrgan} dataset, without performing any downsampling. For validation and testing, we employ the corresponding splits from the DIV2K~\cite{timofte2017ntire} dataset. The input source domain images $\mathcal{X}_\text{val}$ and $\mathcal{X}_\text{test}$ are obtained by first downscaling the images followed by the degradation. For track 1 (see below) we generate ground truth images by applying the degradation directly to the corresponding high-resolution images $\mathcal{Y}_\text{val}^\text{tr1}$ and $\mathcal{Y}_\text{test}^\text{tr1}$. For track 2, users are provided with an additional set of clean high-quality images $\mathcal{Y}_\text{train} = \{y_j\}$ that defines the desired target domain quality. We use the training split of the DIV2K for this purpose to avoid any overlap with the source image set $\mathcal{X}_\text{train}$. For testing and validations we use the same input images and take the unprocessed DIV2K HR images ($\mathcal{Y}_\text{val}^\text{tr2}$ and $\mathcal{Y}_\text{test}^\text{tr2}$) as ground-truth. Visual example input and ground truth images for both tracks are provided in figure~\ref{fig:intro}.

\subsection{Tracks and Competition}

The challenge contains two tracks. Both tracks use an upscaling factor of $4\times$. The competition was organized using the Codalab platform.

\parsection{Track 1: Source Domain}
The aim of this track is to learn a model to capable of super-resolving source domain images (\ie of degraded quality), without changing their low-level image characteristics. That is, the resulting high-resolution images should also be of the source domain quality. Thus, the HR ground truth images $\mathcal{Y}_\text{val}^\text{tr1}$, $\mathcal{Y}_\text{test}^\text{tr1}$ are constructed using the same degradation operation applied directly on the HR image. Participants are only provided the source set $\mathcal{X}_\text{train}$ for training.

\parsection{Track 2: Target Domain}
Here the task is to super-resolve the same source domain images as in Track~1. The difference is that the learned model should generate clean high-quality HR images. The participants are therefore also provided with a second training set $\mathcal{Y}_\text{train}$ defining the target domain quality. This set has no overlap with the source domain set $\mathcal{X}_\text{train}$. The SR method is evaluated against ground truth images of target domain quality ($\mathcal{Y}_\text{val}^\text{tr2}$ and $\mathcal{Y}_\text{test}^\text{tr2}$).

\parsection{Challenge phases} The challenge had three phases: (1) Development phase: the
participants got training images and the LR images of the validation set. (2) Validation phase: the participants had the opportunity to measure performance using PSNR and SSIM metric by submitting their results on the server. A validation leaderboard was also available. (3) Final test phase: the participants got access to the LR test images and had to submit their super-resolved images along with description, code and model weights for their methods.

\parsection{Evaluation protocol} As communicated to the participants at the start of the challenge, the final ranking was decided based on a human perceptual study. The Peak Signal-to-Noise Ratio (PSNR) and the Structural Similarity index (SSIM) \cite{wang2004ssim} was provided on the Codalab platform for quantitative feedback, and also reported in this study. Moreover, we report the LPIPS \cite{zhang2018unreasonable} distance, which is a learned reference-based image quality metric. To obtain a final ranking of the methods, we performed a user study to calculate a Mean Opinion Score (MOS). The test candidates were shown a side-by-side comparison of a sample result and the corresponding ground-truth. They were then asked to evaluate the difference of the two images on a 5-level scale defined as: 0 - 'the same', 1 - 'very similar', 2 - 'similar', 3 - 'not similar' and 4 - 'different'.

\section{Challenge Results}

From 87 registered participants on Track 1, 7 teams entered the final phase and submitted results, code/executables, and factsheets. In Track 2, 4 teams of 75 entered the final phase. All of these teams also participated in Track 1. Only two teams contributed with different solutions in Track 1 and 2. Table~\ref{tab:tr1-test-results}~and~\ref{tab:tr2-test-results} report the final results of Track 1 and 2 respectively, on the test data of the challenge. The Methods of the teams that entered the final phase are described in Section~\ref{sec:methods} and the team's affiliation is shown in Section~\ref{sec:affiliation}.

\subsection{Architectures and main ideas} 
Most teams adopted the ESRGAN generator network or similar ideas for their SR network. The participants presented different innovative solutions to handle the real-world SR setting. The top-ranked MadDemon team aimed to first map the low-resolution training images to the distribution of the source (input) images. This is performed by learning a network that can simulate the the natural image characteristics (i.e.\ degradations) by adding them to bicubically downsampled images. The teams Nam and CVML employ the inverse strategy, i.e.\ to learn a network that first \emph{cleans} the image before super-resolution. The team Image Specific NN for RWSR added used synthetic noise to increase the robustness of their method. The latter approach is also the only team that does not exploit any training data. The networks are learned solely on the input image by adopting the ZSSR~\cite{shocher2018zero} approach. A summary of all participants is given in table~\ref{tab:teams}.

\begin{table}[t]
	\centering
	\newcommand{\sep}{~~}
	\resizebox{\columnwidth}{!}
	{
		\begin{tabular}{@{}ll@{\sep}|@{\sep}c@{\sep}c@{\sep}c@{\sep}c@{}}
			& Method        & $\uparrow$PSNR & $\uparrow$SSIM & $\downarrow$LPIPS  & $\downarrow$MOS\\
			\hline
			\multirow{7}{1mm}{\rotatebox{90}{\resizebox{14mm}{!}{Participants}}}
			& MadDemon                   & 22.65 & 0.48 & 0.36 & \textbf{2.22} \\
			& IPCV IITM                  & 25.15 & 0.60 & 0.66 &         2.36 \\
			& Nam                        & 25.52 & 0.63 & 0.65 &         2.46 \\
			& CVML                       & 24.59 & 0.62 & 0.61 &         2.47 \\
			& ACVLab-NPUST               & 24.77 & 0.61 & 0.60 &         2.49 \\
			& SeeCout                    & 25.30 & 0.61 & 0.74 &         2.50 \\
			& Image Specific NN for RWSR & 24.31 & 0.60 & 0.69 &         2.56 \\
			\hline
			& ESRGAN Superv. & 24.22 & 0.55 & 0.30 & 1.81 \\
			\hline
			\multirow{4}{1mm}{\rotatebox{90}{\resizebox{12mm}{!}{Baselines}}}
			& Bicubic         & 25.34 & 0.61 & 0.73 & 2.43 \\
            & EDSR PT         & 25.14 & 0.60 & 0.67 &  \\
            & ESRGAN PT       & 22.57 & 0.51 & 0.58 &  \\
            & ESRGAN FT-Src   & 24.54 & 0.56 & 0.45 &  \\
			\hline
		\end{tabular}
	}
	\vspace*{0mm}
	\caption{Challenge results for Track 1: Source domain on the final test set.}
	\label{tab:tr1-test-results}
	\vspace*{0mm}
\end{table}

\begin{table}[t]
	\centering
	\newcommand{\sep}{~~}
	\resizebox{\columnwidth}{!}
	{
		\begin{tabular}{@{}ll@{\sep}|@{\sep}c@{\sep}c@{\sep}c@{\sep}c@{}}
			& Method        & $\uparrow$PSNR & $\uparrow$SSIM & $\downarrow$LPIPS  & $\downarrow$MOS\\
			\hline
			\multirow{4}{1mm}{\rotatebox{90}{\resizebox{14mm}{!}{Participants}}}
			& MadDemon                   & 20.72 & 0.52 & 0.40 & \textbf{2.34} \\
			& SeeCout                    & 21.76 & 0.61 & 0.38 &         2.43 \\
			& IPCV IITM                  & 22.37 & 0.62 & 0.59 &         2.51 \\
			& Image Specific NN for RWSR & 21.97 & 0.62 & 0.61 &         2.59 \\
			\hline
			& ESRGAN Superv. & 22.80 & 0.65 & 0.29 & 1.97 \\
			\hline
			\multirow{5}{1mm}{\rotatebox{90}{\resizebox{12mm}{!}{Baselines}}}
            & Bicubic          & 22.37 & 0.63 & 0.66 & 2.60 \\
            & EDSR PT          & 22.35 & 0.62 & 0.60 & \\
            & ESRGAN PT        & 20.76 & 0.52 & 0.55 & \\
            & ESRGAN FT-Src    & 21.36 & 0.54 & 0.53 & \\
            & ESRGAN FT-Tg     & 21.94 & 0.59 & 0.51 & \\
			\hline
		\end{tabular}
	}
	\vspace*{0mm}
	\caption{Challenge results for Track 2: Target domain on the final test set.}
	\label{tab:tr2-test-results}
	\vspace*{0mm}
\end{table}

\begin{figure*}[t]
\centering
	\newcommand{\wid}{.195\linewidth}
	\newcommand{\im}[1]{\includegraphics[width=\wid]{figures/visuals/#1_track_1_0943_q97.jpg}}
	
    \im{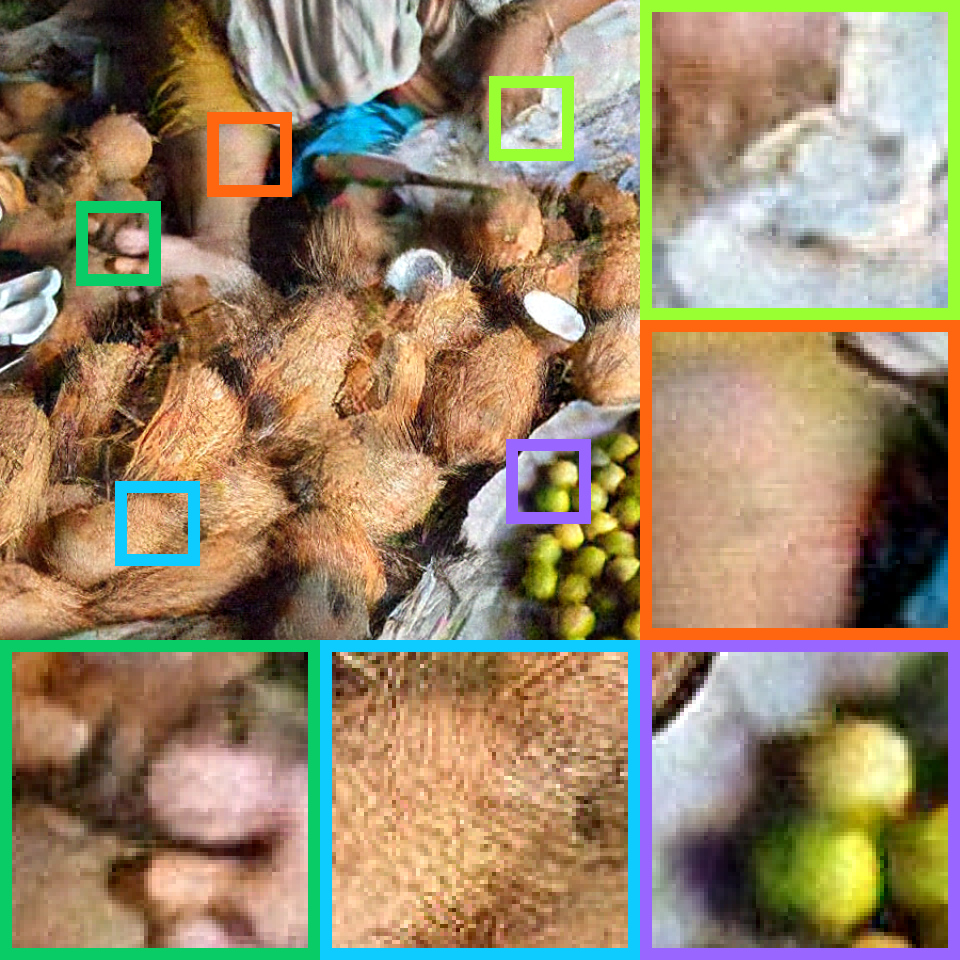}
    \im{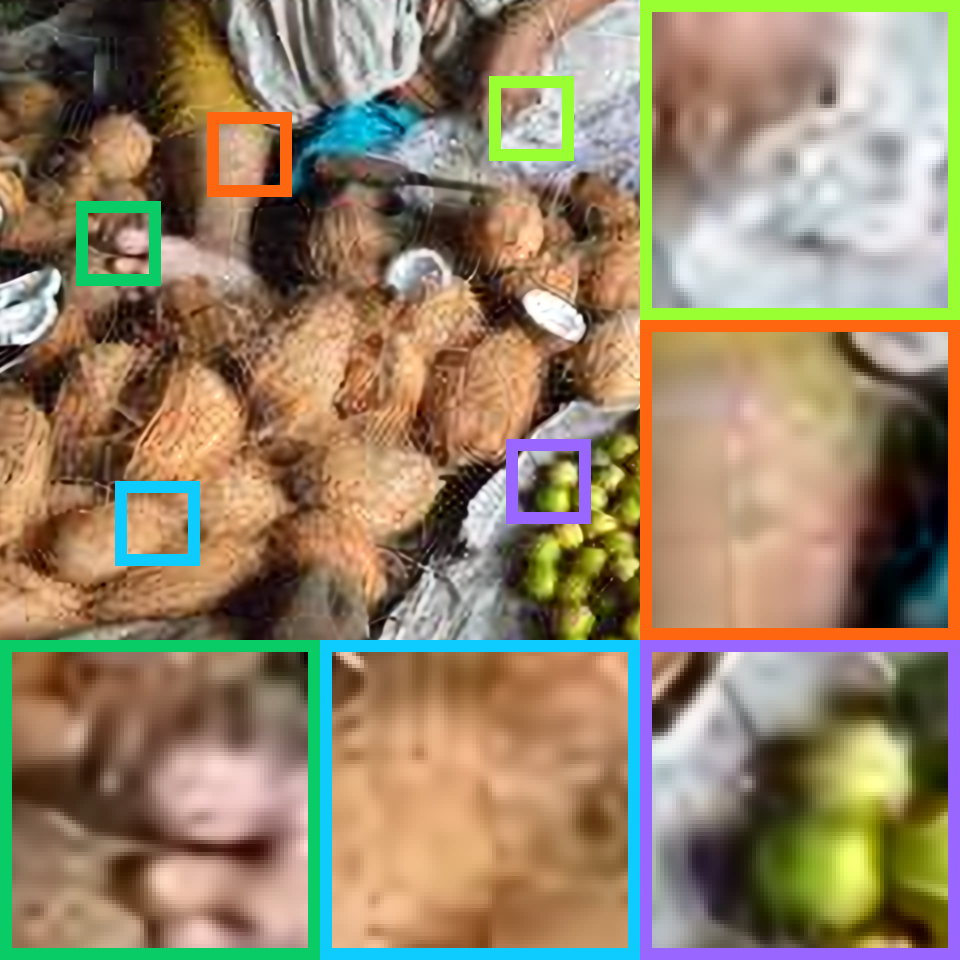} 
    \im{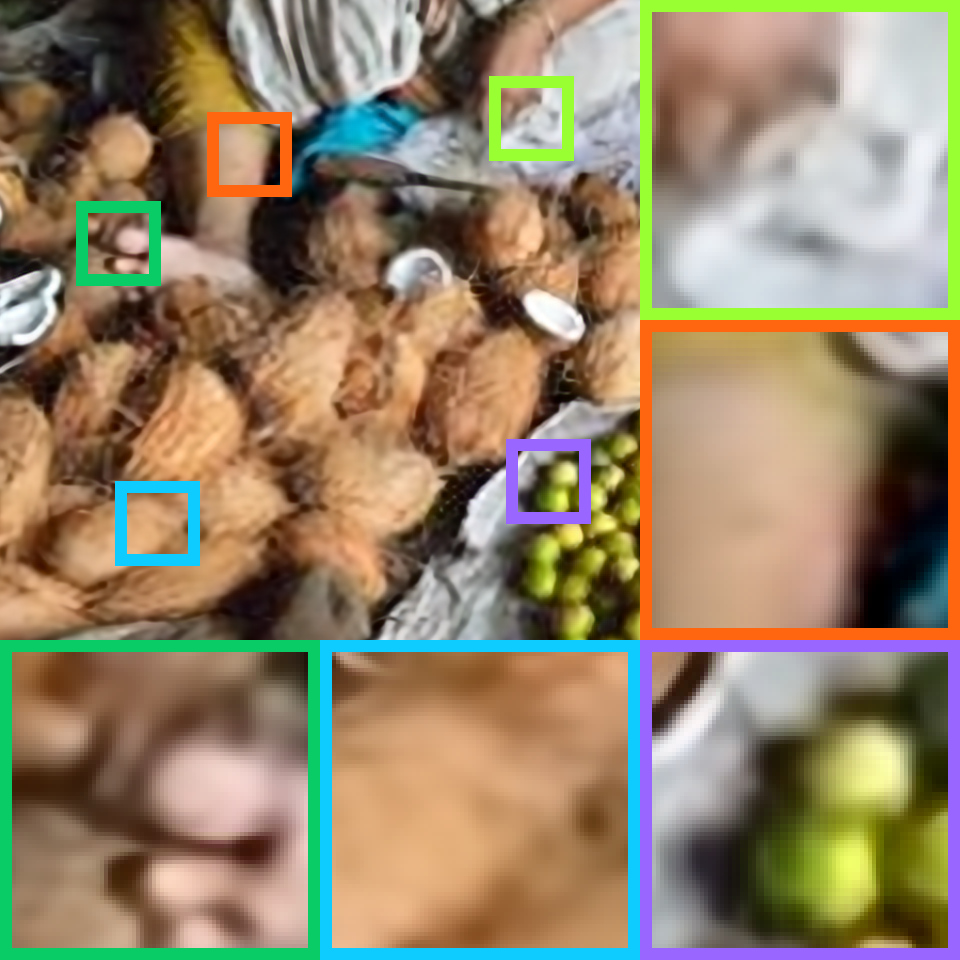}
    \im{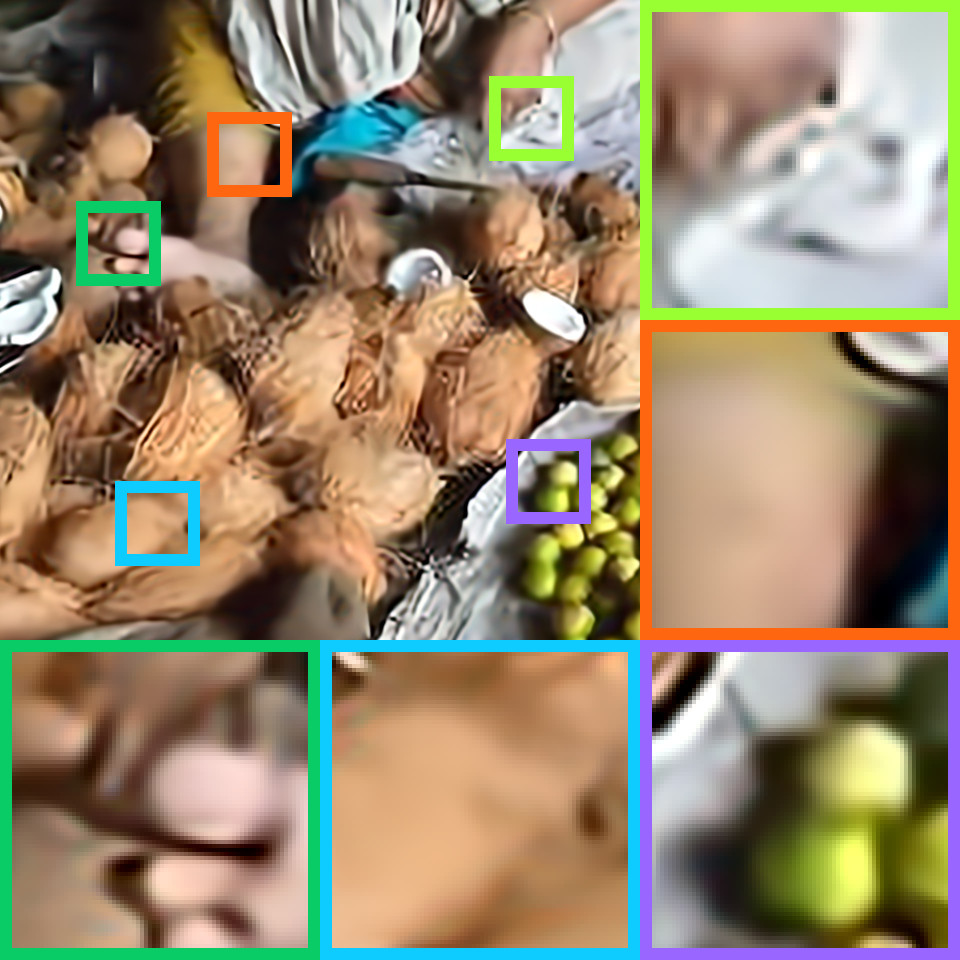}
    \im{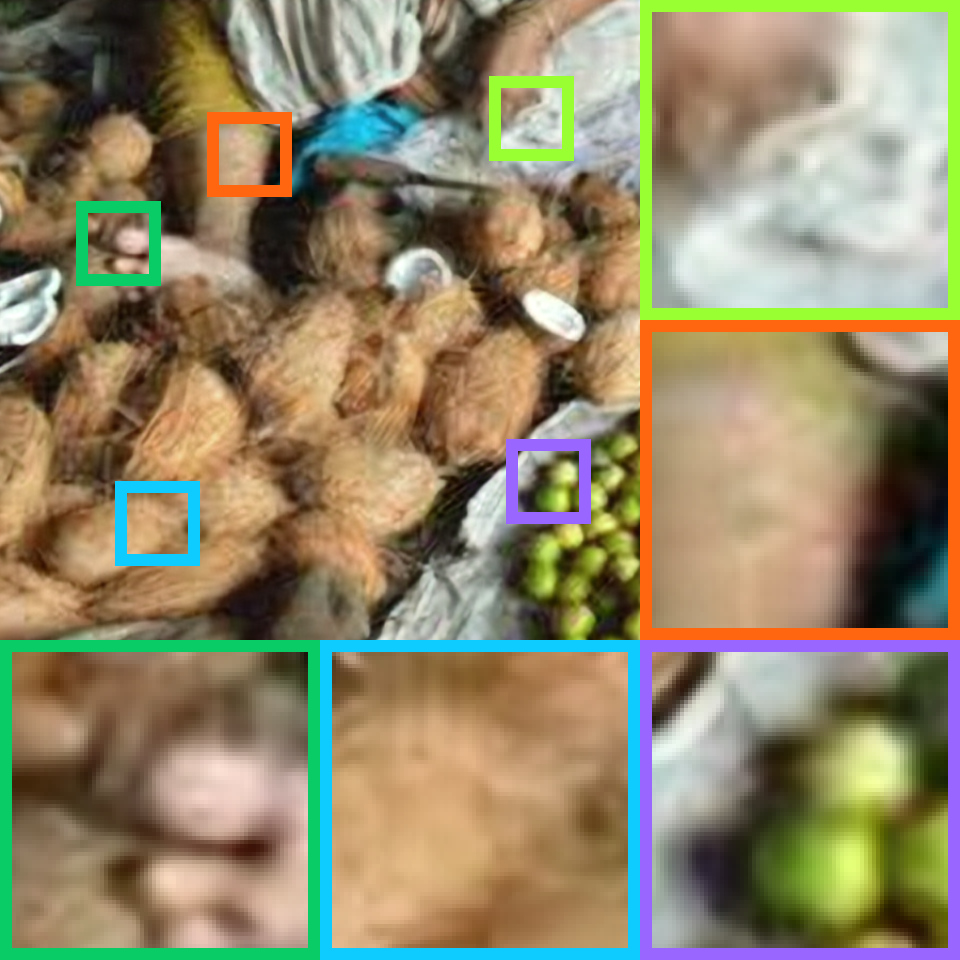}\vspace{-1mm}
	\resizebox{\linewidth}{!}{
		\begin{tabular}{C{4cm} C{4cm} C{4cm} C{4cm} C{4cm}}
			MadDemon & IPCV IITM & Nam & CVML & ACVLab-NPUST
		\end{tabular}
	}
	
	\vspace{1mm}
    \im{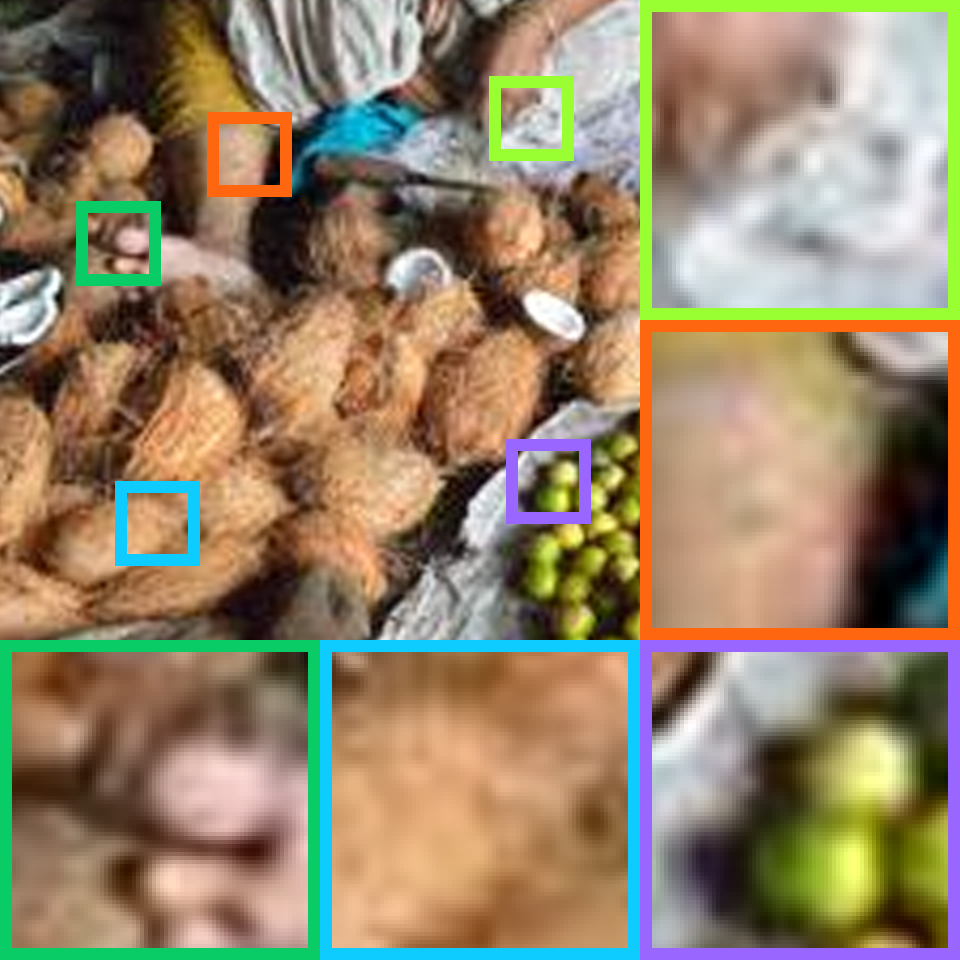}
    \im{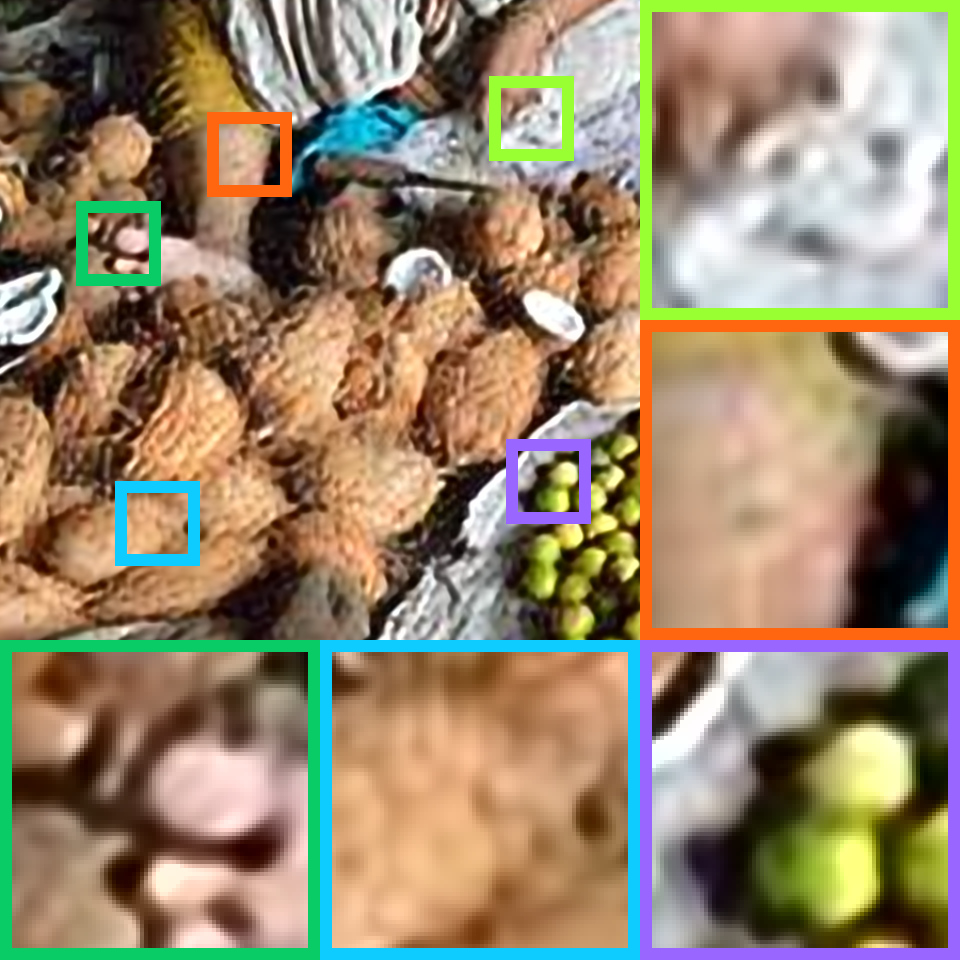}
    \im{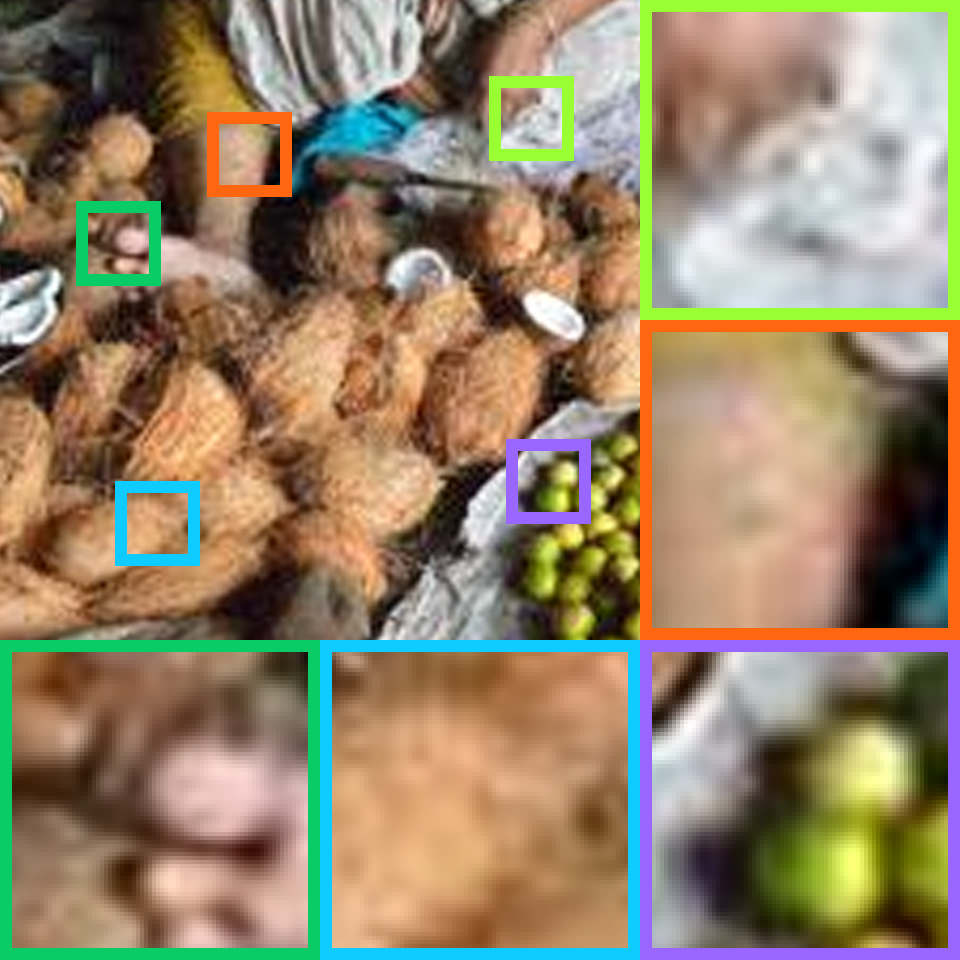}
    \im{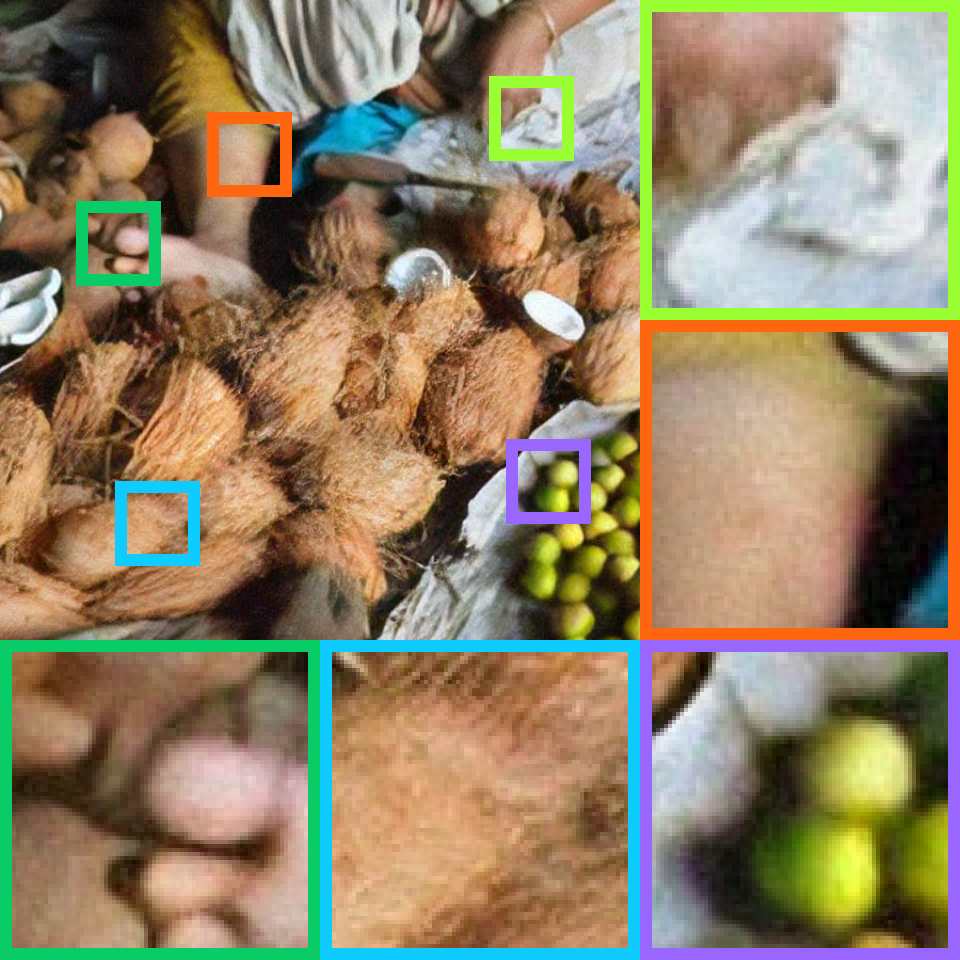}
    \im{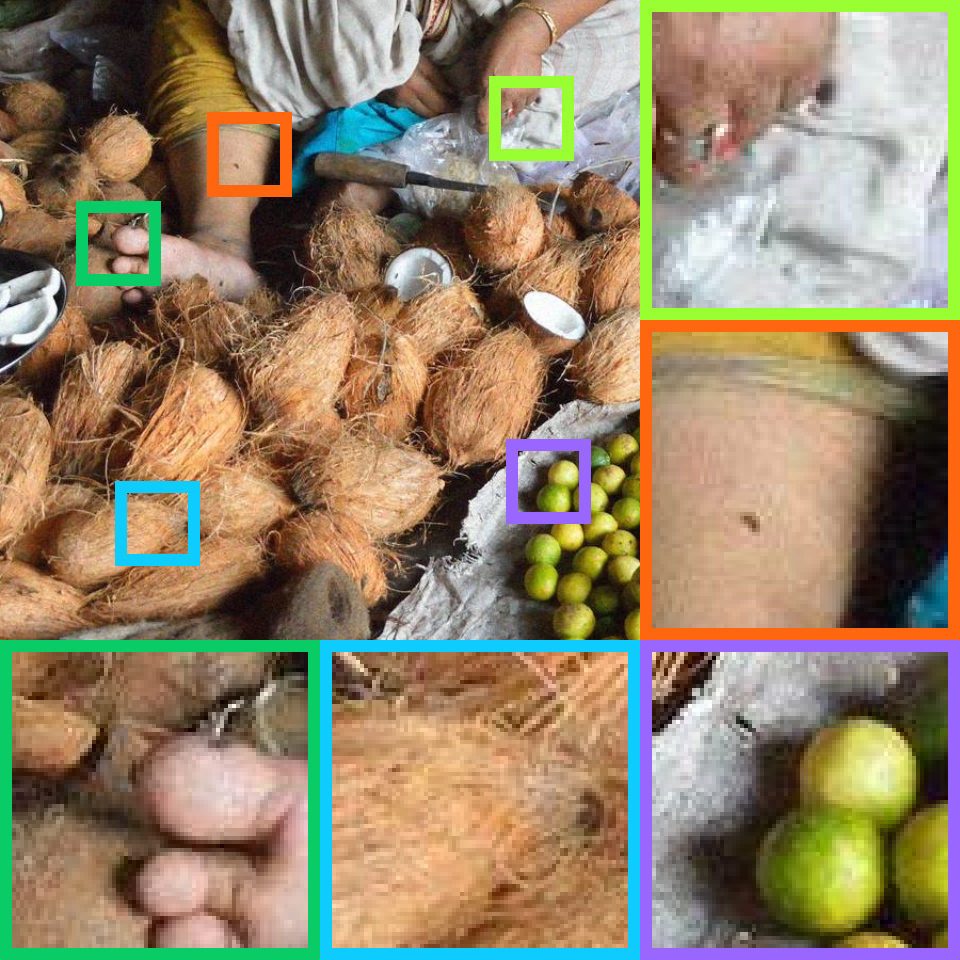}\vspace{-1mm}
	\resizebox{\linewidth}{!}{
		\begin{tabular}{C{4cm} C{4cm} C{4cm} C{4cm} C{4cm}}
            SeeCout & Image Spec.\ NN for RWSR & Bicubic & ESRGAN Superv. & Ground Truth
		\end{tabular}
	}%
	\caption{Qualitative comparison between the challenge approaches for Track 1.}
	\vspace{0mm}
	\label{fig:visuals_track_1}
\end{figure*}

\begin{figure*}[t]
\centering
	\newcommand{\wid}{.16\linewidth}
	\newcommand{\im}[1]{\includegraphics[width=\wid]{figures/visuals/#1_track_2_0943_q97.jpg}}
 
    \im{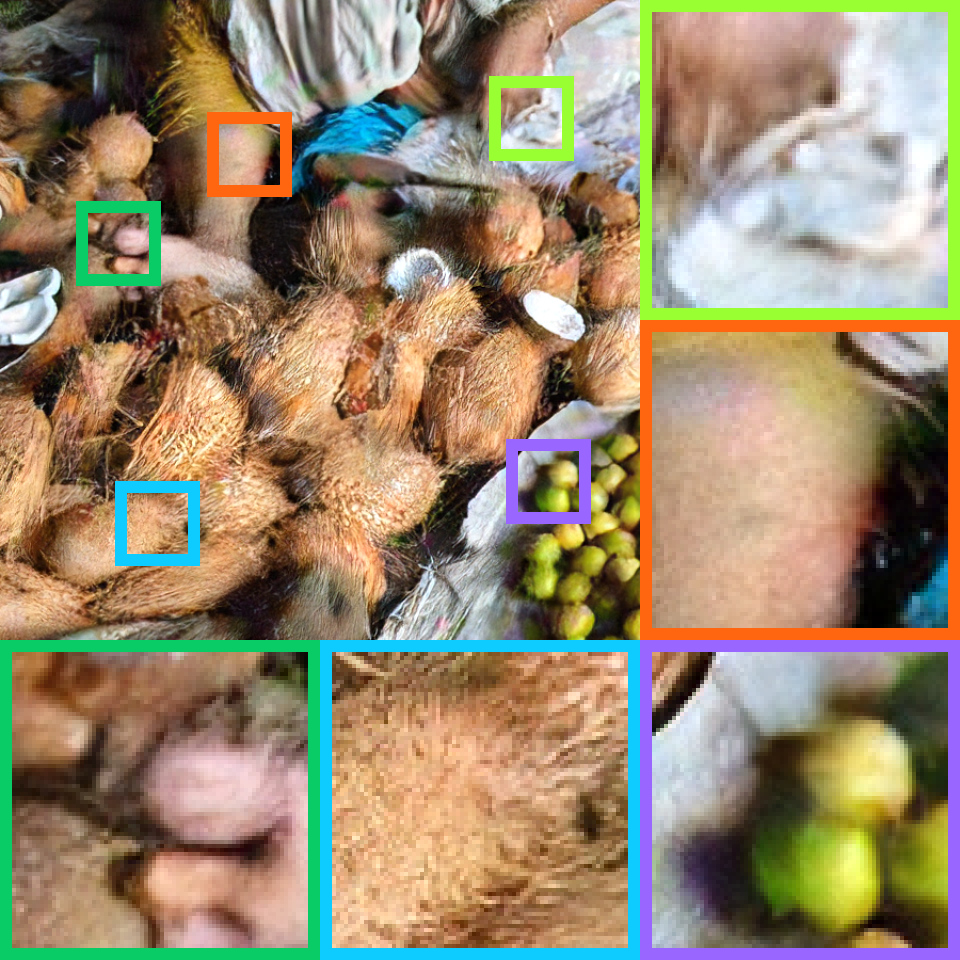}
    \im{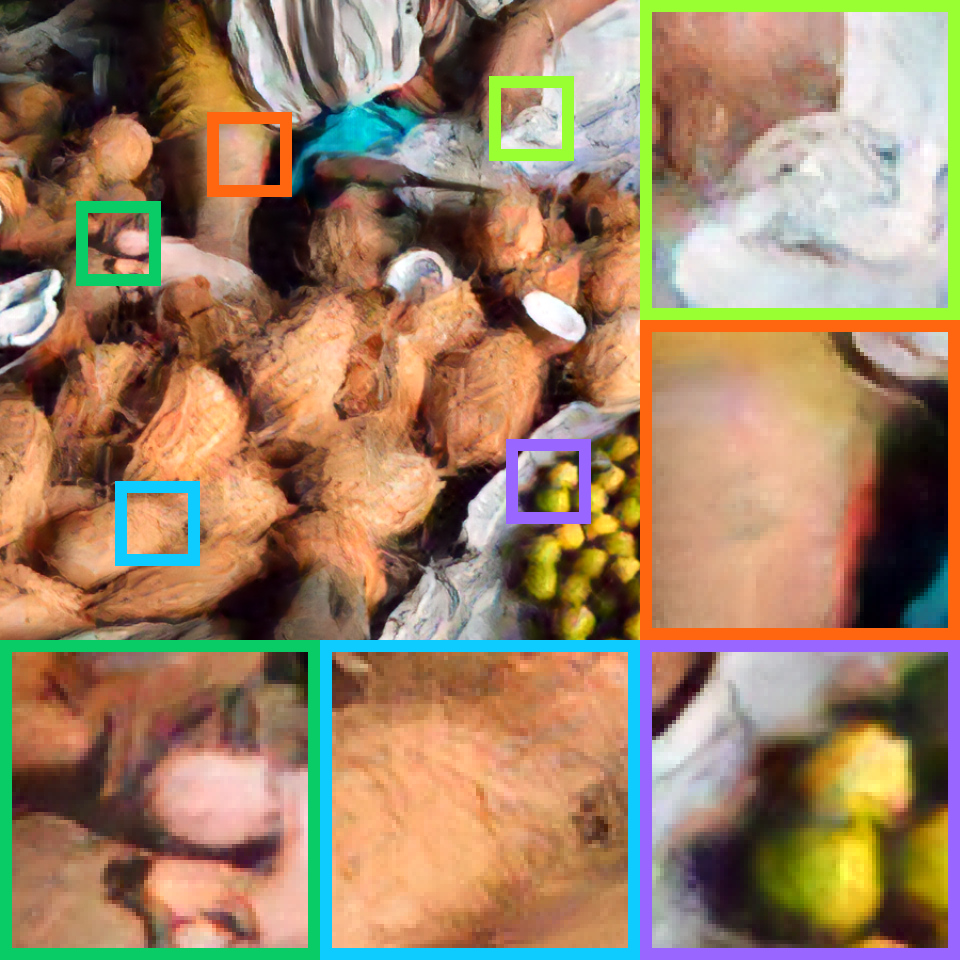}
    \im{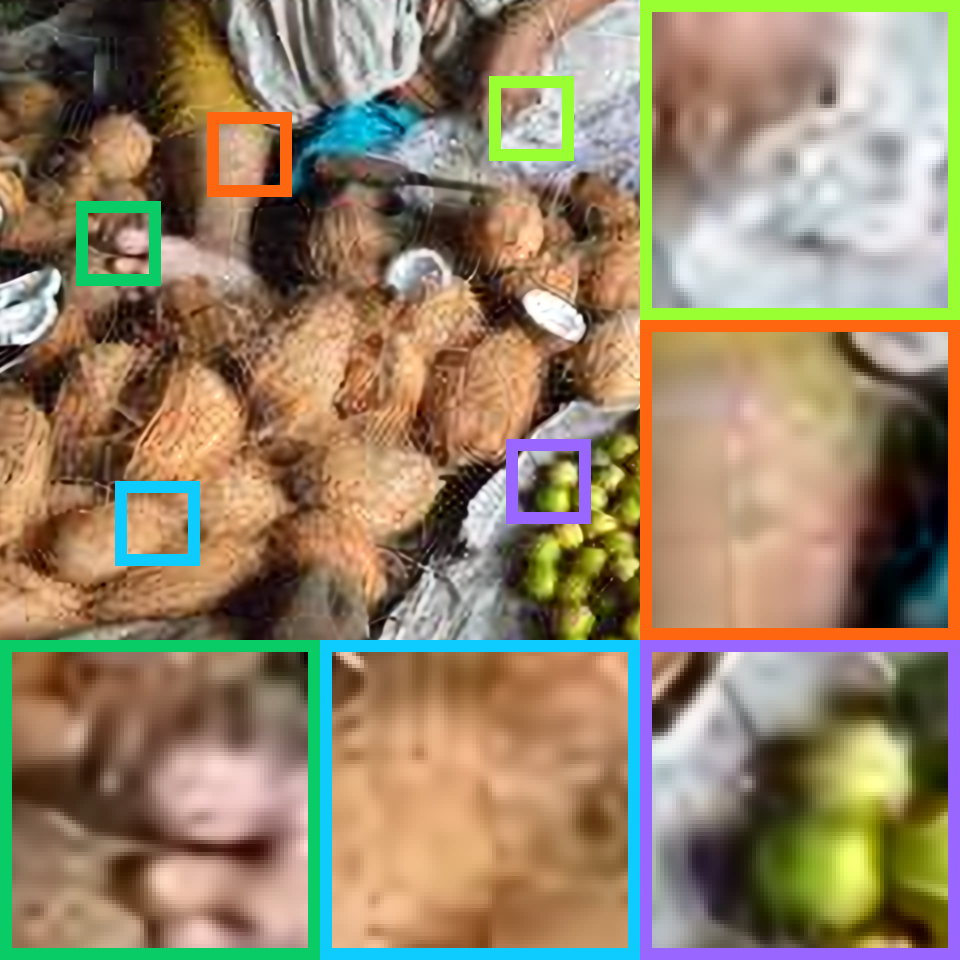}
    \im{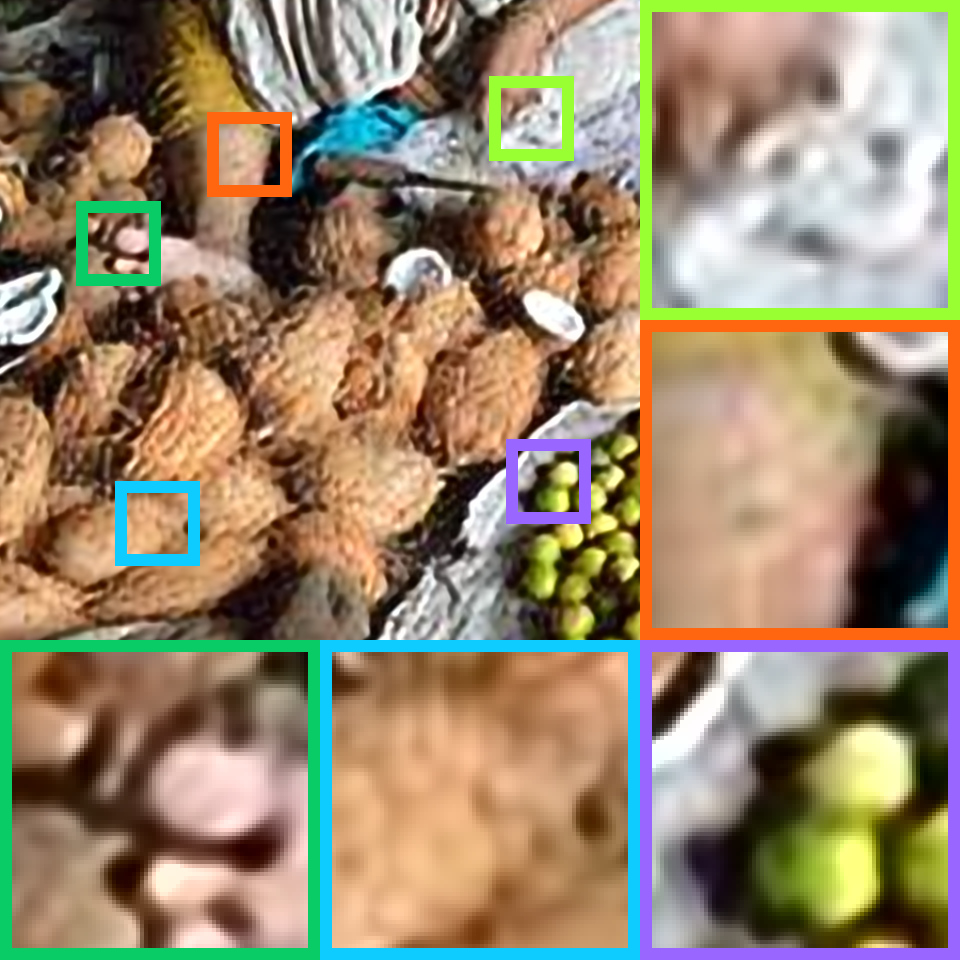}
    \im{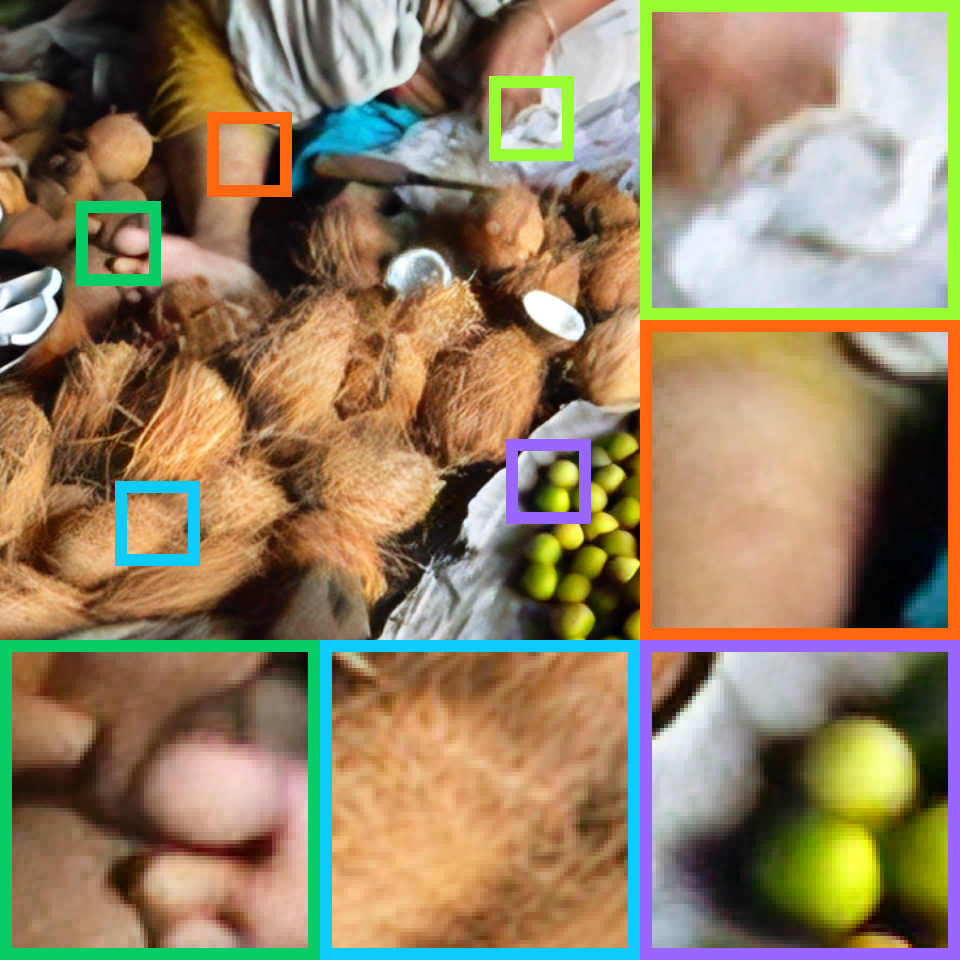}
    \im{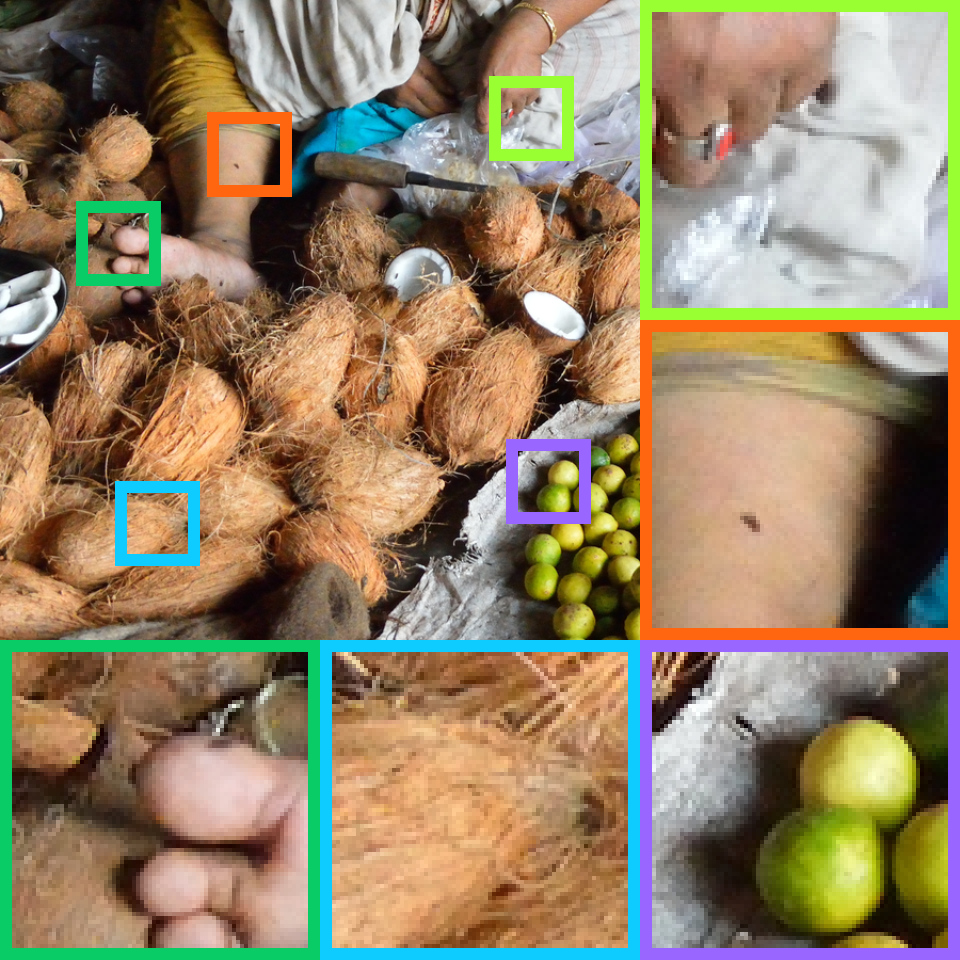}\vspace{-1mm}
	\resizebox{\linewidth}{!}{
		\begin{tabular}{C{4cm} C{4cm} C{4cm} C{4cm} C{4cm} C{4cm}}
            MadDemon &  SeeCout &  IPCV IITM &  Image Spec.\ NN for RWSR & ESRGAN Superv. & Ground Truth
		\end{tabular}
	}%
	\caption{Qualitative comparison between the challenge approaches for Track 2. The bicubic interpolation is the same as for track 1.}\vspace{-3mm}
	\label{fig:visuals_track_2}
\end{figure*}

\subsection{Baselines}
We compare methods participating in the challenge with several baseline approaches.

\parsection{Bicubic}
Standard bicubic upsampling using MATLAB's imresize function.

\parsection{EDSR PT}
The pre-trained EDSR \cite{lim2017EDSR} method, using the network weights provided by the authors. The network was trained with clean images using bicubic down-sampling for supervision.

\parsection{ESRGAN PT}
The pre-trained ESRGAN \cite{wang2018esrgan} method, using the network weights provided by the authors. The network was trained with clean images using bicubic down-sampling for supervision. Unlike EDSR, it includes a perceptual GAN loss.

\parsection{ESRGAN FT-Src}
An ESRGAN network that is fine-tuned on the source domain training set $\mathcal{X}_\text{train}$ for the challenge. This is performed using simple bicubic downsampling as supervision. The network is initialized with the pre-trained weights provided by the authors.

\parsection{ESRGAN FT-Tg}
An ESRGAN network that is fine-tuned only on the target domain training set $\mathcal{Y}_\text{train}$ for the challenge. This is performed using simple bicubic downsampling as supervision. The network is initialized with the pre-trained weights provided by the authors.

\parsection{ESRGAN Superv.}
An ESRGAN network that is fine-tuned in a fully supervised manner, by applying the synthetic degradation operation used in the challenge. The degradation was unknown for the participants. This method therefore serves as an upper bound in performance, allowing us to analyze the gap between supervised and unsupervised methods. For Track 1 and Track 2 we employ the source $\mathcal{x}_\text{train}$ and target $\mathcal{Y}_\text{train}$ domain train images respectively. Low-resolution training samples are constructed by first down-sampling the image using the bicubic method and then apply the synthetic degradation. The network is thus trained with real input and output data, which is otherwise inaccessible. As for previous baselines, the network is initialized with the pre-trained weights provided by the authors.

\subsection{Results}

Here we present the final results of the challenge. All experiments presented were conducted on the test set for the respective tracks.

\parsection{Human perceptual study} In Track 1 (table~\ref{tab:tr1-test-results}) the team MadDemon achieves the best MOS score, with an $0.21$ improvement over Bicubic upsampling. The second-best score is achieved by the IPCV\_IITM team, which is $0.07$ better than Bicubic upsampling. For all other competing approaches, the MOS is surprisingly slightly worse than that of the Bicubic method. We attribute this to the challenging conditions posed in the competition. As verified by the visual results (see figure~\ref{fig:visuals_track_1} and \ref{fig:visuals_track_2}), deep SR approaches are sensitive to such high-frequency degradations, unless explicitly trained to handle them.

For Track2 (table~\ref{tab:tr2-test-results}), the MadDemon team also achieves the best MOS, with $0.26$ better than Bicubic. SeeCout achieves the second best, with an $0.17$ improvement over Bicubic in MOS. These were the only two teams that submitted specialized solutions for Track 2. Contrary to Track 1 however, all participating teams achieve a MOS score better than that of the Bicubic method. 

Interestingly, for both tracks there is a substantial margin between the best participating methods and ERSRGAN model that was trained with full supervision (i.e.\ with access to the used degradation model). This indicates that there is still large scope for future research in developing and improving unsupervised learning techniques for super-resolution.

\parsection{Computed Metrics}
Already at the start of the challenge, it was clarified that the final evaluation will be performed by a human study. We report PSNR and SSIM for reference. From the results it is clear that the participating methods did not optimize for fidelity only. In fact, Bicubic achieves the second best PSNR for track 1 and the best (together with IPCV IITM) for track 2. It is noteworthy that the best performing method in terms of MOS (MadDemon) provides significantly lower PSNR and SSIM scores compared to all other participants. This suggest that the MadDemon team strongly focused on perceptual quality. It also confirms the limitations of the PSNR and SSIM metrics for perceptual evaluation, that were brought up by~\cite{ledig2017photo}. However, it should also be noted that the ESRGAN supervised model achieves a superior MOS while also providing a very competitive PSNR and SSIM.

We also report the learned LPIPS distance \cite{zhang2018unreasonable} for all approaches. It is a reference-based image quality metric, computed as the $L^2$ distance in a deep feature space, which is fine-tuned to correlate better with human perceptual opinions. However, the interpretation of these results is complicated by the fact that the MadDemon team explicitly employed LPIPS as a perceptual loss for training their networks (although with a different LPIPS backbone network). Other methods, such as the ESRGAN-based ones, employ feature-based losses using ImageNet pre-trained VGG networks, which in its design is very similar to LPIPS. Thus, although the LPIPS score seems to better correlate with MOS, it is difficult to draw any clear conclusions from this result. For Track 1, MadDemon achieves an LPIPS almost that of the ESRGAN supervised, although its MOS is significantly lower. In this track, the ESRGAN FT-Src achieves a competitive LPIPS, however it was not included for the perceptual study.

\parsection{Qualitative Analysis}
\label{parsec:quality}
Qualitative results are shown in figure~\ref{fig:visuals_track_1} and \ref{fig:visuals_track_2} for Track 1 and 2 respectively.
Of the participating methods in Track 1, MadDemon achieves the visually most pleasing results. The approach is even able to reproduce the $8\times 8$ blocks, stemming from the source degradation, in the high resolution. Interestingly, such characteristics is not even present in the ESRGAN supervised model results. This is likely due to the texture loss employed by the MadDemon team. In the results of the IPCV\_IITM, ACVLab-NPUST, SeeCout and Image Spec.\ NN for RWSR teams, one can clearly observe $32\times 32$ blocks artifacts, stemming from the upscaling of the $8\times 8$ blocks in the source input data. The cleaning-based approach of Nam and CVML successfully alleviates the block artifacts, but over-smooths the images instead.

Also in Track 2, the MadDemon team provides the visually most pleasing results, comparable to the ESRGAN supervised model on the given example in figure~\ref{fig:visuals_track_2}. The SeeCout results suffer from a slight color shift and more cartoon-looking results. The other participants generate more severe block and high-frequency artifacts.
Regarding the overall texture quality, the results of MadDemon have a crisp appearance with rich high frequency components. Despite that, they also have a typical GAN appearance, leading to a slight painting-like appearance. The textures of the results of Image Specific NN for RWSR are much less crisp, but still impressive considering that the model is only trained on the input image.

\section{Challenge Methods and Teams}
\label{sec:methods}

This sections give brief descriptions of the participating methods. A summary of all participants is given in table~\ref{tab:teams}.

\begin{table*}[t]
	\centering
	\newcommand{\sep}{~~}
	\resizebox{\textwidth}{!}
	{
		\begin{tabular}{@{}ll@{\sep}|@{\sep}c@{\sep}c@{\sep}c@{\sep}c@{\sep}c@{}}
        Team & User & Track 1 & Track 2 & Additional Data & Run time Track 1 & Run time Track 2 \\
        \hline
        SeeCout                    & SeeCout         & yes   & yes   & None                & 1.62s & 1.76s \\
        Image Specific NN for RWSR & sefi            & yes   & yes   & None                & 360s  & 360s  \\
        IPCV IITM                  & kuldeeppurohit3 & yes   & yes   & N/A                 & 15s   & 15s   \\
        Nam                        & Nam             & yes   & no    & None                & 12    & N/A   \\
        CVML                       & DokyeongKwon    & yes   & no    & Images from Track 2 & 0.43s & N/A   \\
        ACVLab-NPUST               & jesse1029       & yes   & no    & N/A                 & N/A   & N/A   \\
        MadDemon                   & MadDemon        & yes   & yes   & None                & 1.5s  & 1.5s  \\
        \hline
		\end{tabular}
	}
	\vspace*{0mm}
	\caption{Challenge Participants and methods.}
	\label{tab:teams}
	\vspace*{0mm}
\end{table*}

\begin{figure}[t]
	\centering
	\includegraphics[width=\columnwidth]{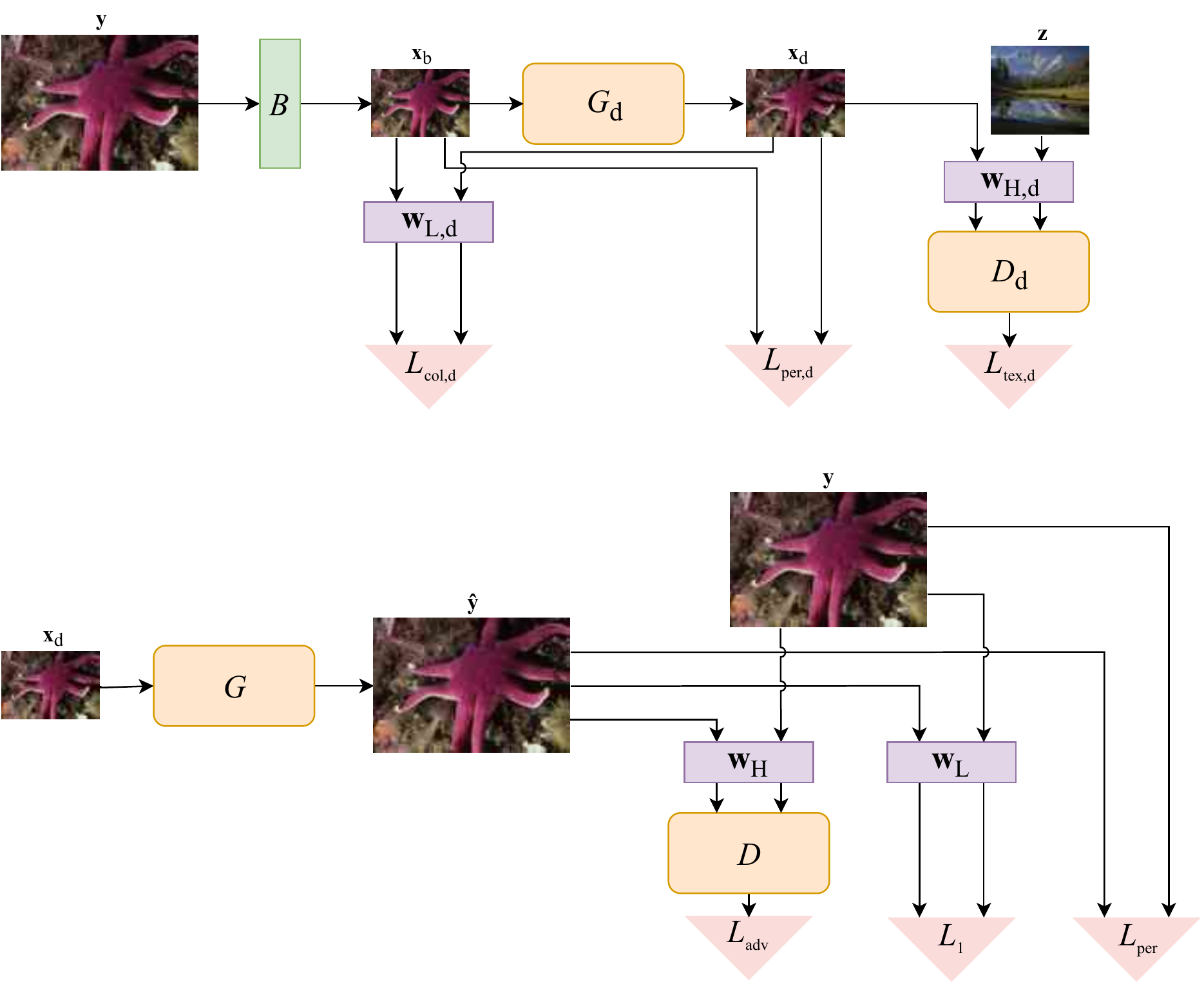}
	\caption{The overall architecture of MadDemon team.}
	\label{fig:maddemon}
\end{figure}
\begin{figure}[t]
\centering
\includegraphics[width=\columnwidth]{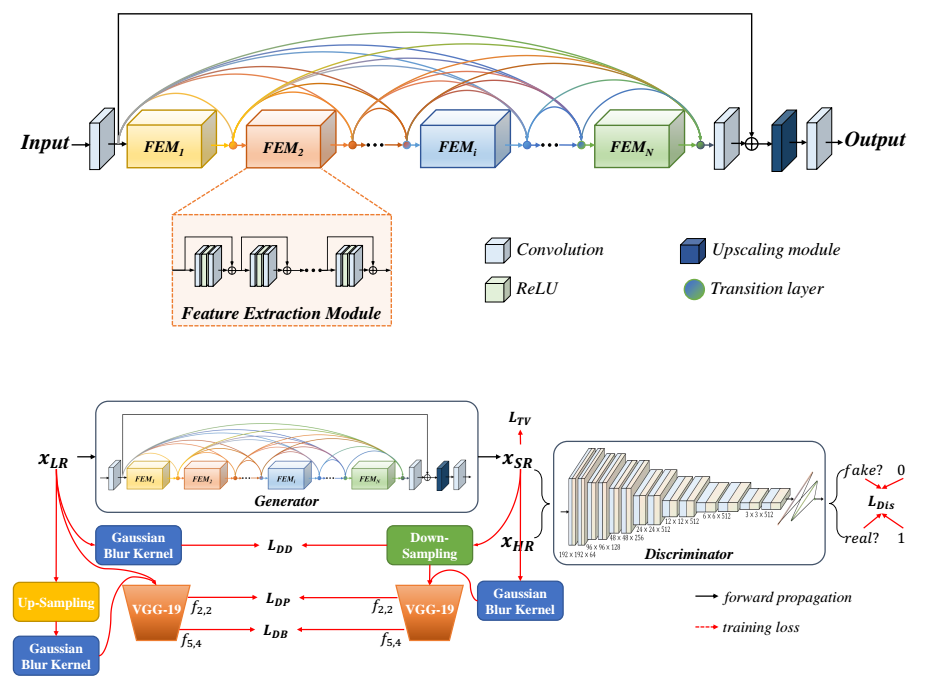}
\caption{The overall architecture of SeeCout team.}
\label{fig:seecout}
\end{figure}

\parsection{MadDemon team}
introduce a neural network model called DSGAN, which allows them to downscale images while also keeping their natural image characteristics. DSGAN can be trained in an unsupervised fashion on the original images to generate image pairs that
have the same kind of corruptions. The team then uses the generated data to
train an SR model based on ESRGAN \cite{wang2018esrgan}, which greatly improves its performance on real-world data. Furthermore, they propose to separate the low
and high frequencies and treat them differently during the training of the SR
model. Since the low frequencies are preserved by downsampling operations, its corresponding upsampling operation can be trained using a simple pixel-wise loss.
This means that only the high frequencies require additional adversarial
training, which simplifies the GAN learning. This idea is applied to both the proposed DSGAN model and the used ESRGAN SR model. The image frequency content is separated by simply applying low and high-pass filters before the loss function and before the discriminator. Furthermore they use LPIPS\cite{zhang2018unreasonable} as a loss to increase the perceptual quality.

In detail, the team first train the DSGAN model on the source dataset. The model itself has 8 residual blocks with 64 channels with a corresponding discriminator with 4 conv layers. The DSGAN is then applied to the given source and target data to create the corrupted training data for the SR model. For Track 1 the DSGAN model is applied to the target dataset images, which are used together with the source domain data as the HR dataset. For Track 2 the
original target dataset and the source dataset bicubicly downscaled with
a factor of 2 to clean it up. In both cases, the LR training images are created by first bicubicly downscaling the images and then applying the trained DSGAN
model. Using the created data, a slightly modified ESRGAN model is then finetuned
 for 50k steps. The overall architecture is visualized in figure~\ref{fig:maddemon}.
 
\begin{figure}[t]
	\centering
	\includegraphics[width=\columnwidth]{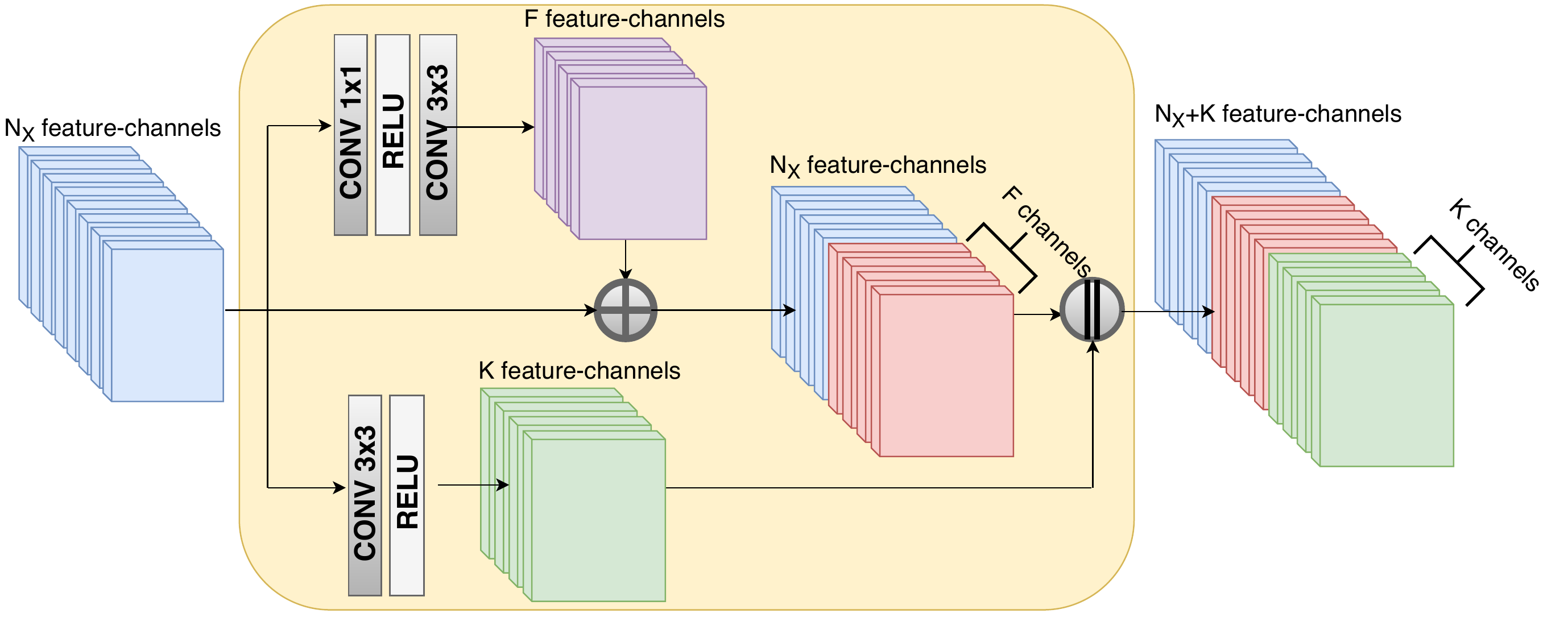} \\ \vspace{2mm}
	\includegraphics[width=\columnwidth]{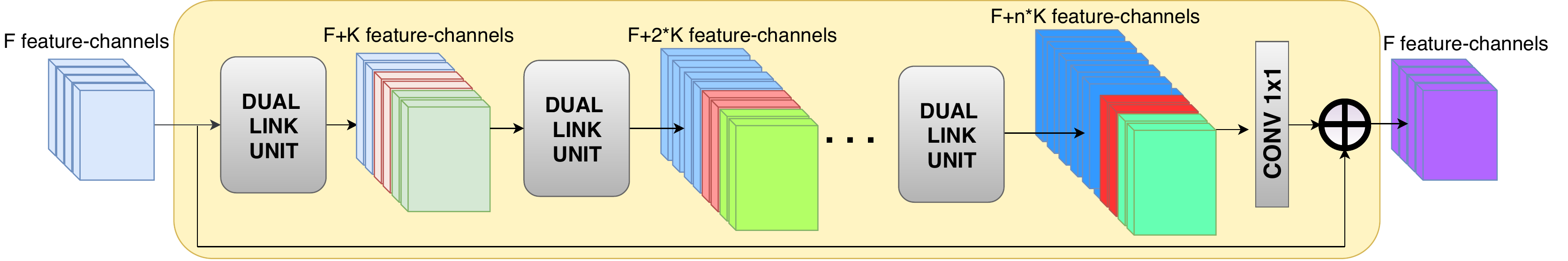} \\ \vspace{2mm}
	\includegraphics[width=\columnwidth]{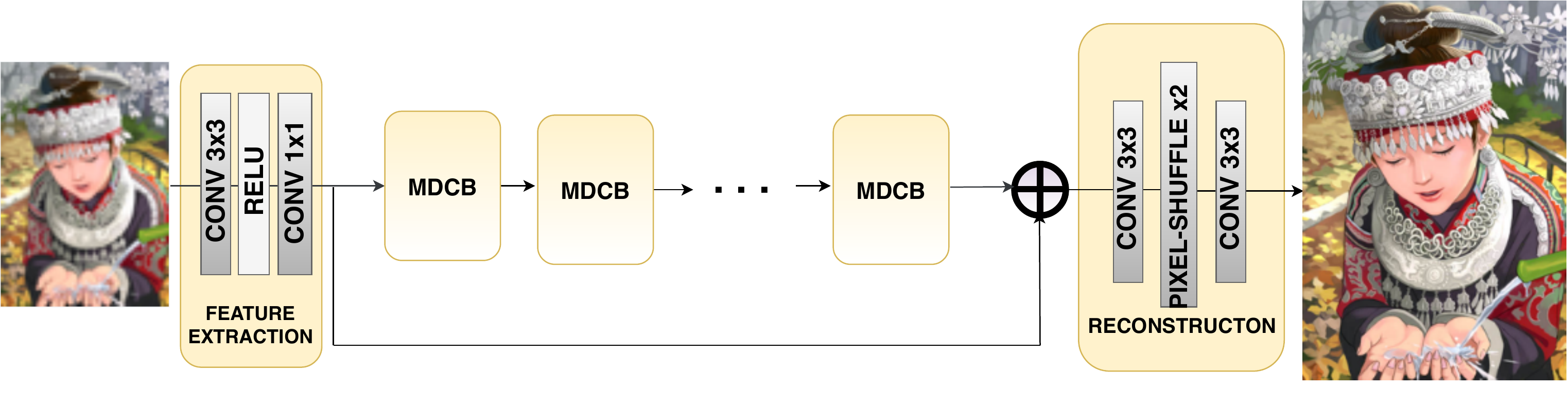}
	\vspace{-3mm}
	\caption{The Dual Link (top), mixed-dense connection block (MDCB) (middle) and the full SR architecture (bottom) of the IPCV\_IITM team.}
	\label{fig:MDCB}
	\vspace{-2mm}
\end{figure}     
 
\parsection{IPCV\_IITM team} proposes the Mixed-Dense Connection Network (MDCN), which employs dense network topologies to design asuper-resolution architecture that benefits from a mixture of such connections. It uses the Mixed-Dense Connection Blocks (MDCBs), which contain a rich set of connections to enable efficient feature-extraction and ease the gradient propagation. In each MDCB, $n$ Dual Link Units are present. Additive links in the unit grant the benefits of reusing common features with low redundancy, while concatenation links give the network more flexibility in learning new features. Each Dual Link Unit performs the additive operation to the last $F$ features of the input and concatenating connections for the rest. This design is an improved version of the network in \cite{purohit2018scale}. A visual depiction of these connections and the MDCB can be seen in the Fig.~\ref{fig:MDCB}. Furthermore, a gating mechanism is employed to allow larger growth rate by reducing the number of features, which stabilizes the training of a wide network. Each convolution or deconvolution layer is followed by a rectified linear unit (ReLU) for nonlinear mapping, except for the final $1\times1$ layer.
The complete network (MDCN) broadly consists three parts: initial feature extraction module, a series of mixed-dense connection block (MDCBs) and an HR reconstruction module (Fig.~\ref{fig:MDCB}). 
The same model was used for both tracks. 
 
\begin{figure}[t]
	\centering
	\includegraphics[width=\columnwidth]{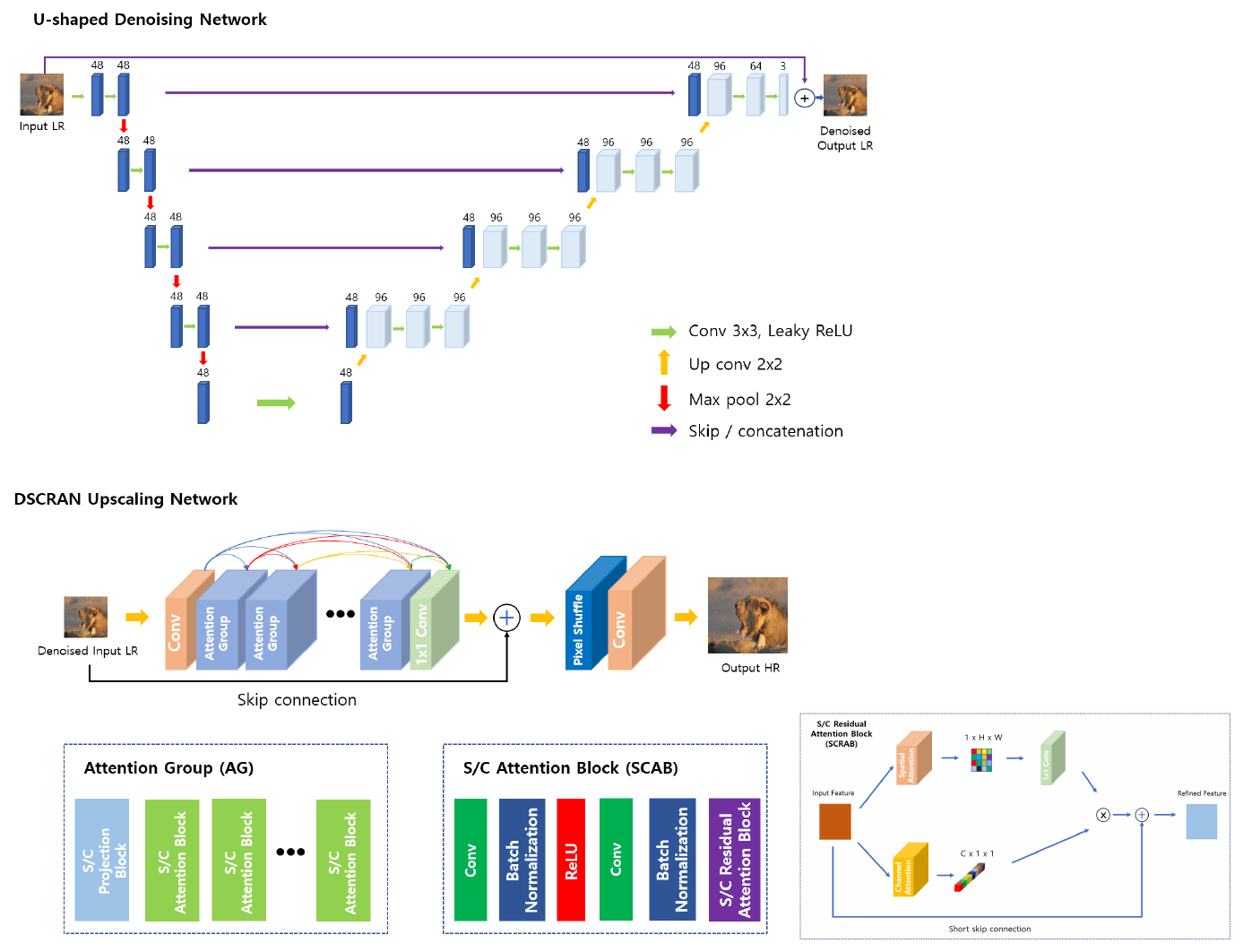}\vspace{-2mm}
	\caption{The cleaning network (top) and SR network with dense connections and attention modules (bottom) of the Nam team.}
	\label{fig:Nam}
	\vspace{-2mm}
\end{figure}

\parsection{Nam team}
uses a two-step approach that first cleans the low-resolution image and then super-resolves it. For cleaning the image they train a U-shaped network that is trained supervised. The training pairs were generated by applying synthetic noise to low-resolution training images. The second network super-resolves the the cleaned image. The team employs a network architecture based on EDSR \cite{lim2017EDSR}, but adding simultaneous channel and spatial attention modules. The team also add dense blocks, similar to ESRGAN \cite{wang2018esrgan}, to prevent the extracted features from vanishing. The network structure is shown in Figure~\ref{fig:Nam}

In detail the cleaning network uses a U-net structure with max-pooling and batch normalization in five scale levels. The latent space has 48 channels and the upsampling path employs 64 channels. The super-resolution network uses five attention groups with one Spacial Channel Projection Block and 35 Spacial Channel Attention blocks.

\begin{figure}[t]
	\centering
	\includegraphics[width=\columnwidth]{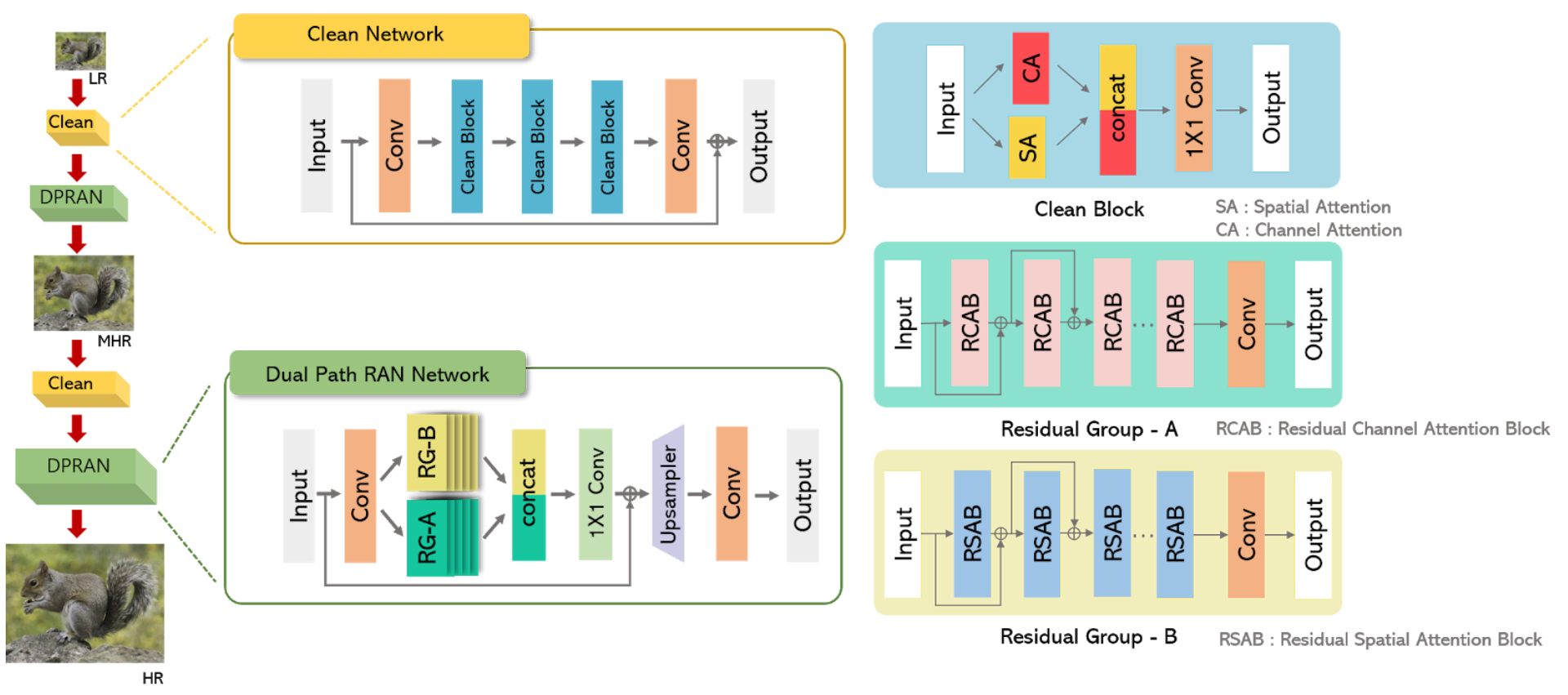}
	\vspace{-2mm}
	\caption{The overall architecture of CVML team.}
	\label{fig:CVML}
	\vspace{-2mm}
\end{figure}

\parsection{CVML team} does the super-resolution in two steps of a factor of 2 each. These steps consist of a cleaning and an upsampling network, as shown in Figure~\ref{fig:CVML}. Both networks are using the method of dual-path residual attention mechanism. The dual-path consists of two parts. The first path is channel attention path and next is the spatial attention path. Their super-resolution model consists of three parts: residual groups, upsampling layer, and convolution layer. They employ two types of residual groups. One for channel attention blocks and the other for spatial attention. They then concatenate the deep feature obtained by passing through these two groups and put the result of passing a $1\times1$ convolution layer into an upsampling layer for upscaling. 

\parsection{ACVLab-NPUST team} fuses a modified ESRGAN and SRResNet as their super-resolution pipeline. The modified ESRGAN discriminator employs the Hswish activation function and uses a global average pooling layer no fully connected layers. In the generator network, more shortcuts are added between Dense-Block connections. A procedure for adaptively controling the weights of the losses is also proposed.
The final result is obtained by fusing the two output images with factor 0.8 for the modified EDSRGAN and 0.2 for SRResnet.

\parsection{SeeCout team}
propose the un-paired super-resolution solution with Degradation
Consistency (DCSR) method. To retain the structures and contents of super-resolved images from
low-resolution inputs, several losses are introduced that imposes consistency between super-resolved and low-resolution images. A down-scaling degradation consistency is added by downsampling the super-resolved image and comparing it with the blurred low-resolution image. The blurring removes much of the degradation in the original LR image. Furthermore, perceptual degredation consistency is enforced by comparing extracted VGG-19 features. One loss is computed in high resolution by taking high-level (5th layer) features from the output image and the bicubically upsampled LR image. Both images are blurred before feature extraction to remove the effect of noise. A similar loss is computed in low resolution between extracted low-level features (2nd layer) from the downscaled SR image and the LR image. As a second contribution, the team propose and efficient generator architecture aggregating multi-level features employing dense connections on multiple deep feature extraction modules. Lastly, for Track 2 the team also employs a relativistic GAN discriminator, similar to ESRGAN \cite{wang2018esrgan}. The overall architecture is visualized in figure~\ref{fig:seecout}.

\parsection{Image Specific NN for RWSR team}
combines two separate works, namely KernelGAN \cite{kernelgan} and ZSSR \cite{shocher2018zero} with modifications for the task. KernelGAN produces the image specific SR kernel estimation and ZSSR produces a SR image w.r.t.\ that kernel. Both these approaches employ fairly shallow CNNs, that are trained solely on the input image (i.e. image-specific), therefore requiring a new training process for every image. 
In the first stage, the image specific SR kernel estimation is found using an internal-GAN. The generator is trained to downscale the input image by applying a \emph{patch discriminator} loss between w.r.t.\ real input image patches. Once trained, the generator constitutes the kernel that best preserves the patch distribution across scales. Next, the input image is downscaled using the kernel estimation from stage 1, to obtain a low resolution image. In the second stage, the ZSSR network is trained to enhance patches from the generated LR image to their the corresponding patches of the input image. Once trained, the ZSSR network can be applied to the full input image to obtain the SR output. To achieve better robustness to degradations, the team adds synthetic noise to the LR images that are used for training the ZSSR network.

	\section{Conclusions}
This paper presents the setup and results of the AIM 2019 challenge on real world super-resolution. Contrary to conventional super-resolution, this challenge addresses the real world setting, where paired true high and low-resolution images are unavailable. For training, only one set of source input images were provided to the participants. The challenge contains two tracks, where the aim was to super-resolve images with preserved characteristics (Track 1) or achieve images of a target quality (Track 2). The challenge had in total 7 teams competing. The participating methods demonstrated interesting and innovative solutions to the real-world super-resolution setting. While the best methods achieved results better than the final baseline, a large margin to fully supervised approaches still remain. Our goal is that this challenge stimulates future research in the area of unsupervised learning for image super-resolution and other similar tasks, by serving as a standard benchmark and by the establishment of new baseline methods.
	\section{Acknowledgements}
We thank the AIM 2019 sponsors.

	\section*{Appendix A: Teams and affiliations}
\label{sec:affiliation}

\subsection*{AIM2019 organizers}
\textbf{Members:} Andreas Lugmayr (andreas.lugmayr@vision.ee.ethz.ch), Martin Danelljan (martin.danelljan@vision.ee.ethz.ch), Radu Timofte (radu.timofte@vision.ee.ethz.ch)

\textbf{Affiliations:} Computer Vision Lab, ETH Zurich

\subsection*{MadDemon team}

\textbf{Title:} Unsupervised Real World Super-Resolution

\textbf{Members:} Manuel Fritsche (manuelf@ethz.ch), Gu Shuhang, Radu Timofte

\textbf{Affiliations:} Computer Vision Lab, ETH Zurich

\subsection*{IPCV IITM team}
\textbf{Title:} Super-Resolution Using Scale-Recurrent Residual Dense Networks

\textbf{Members:} Kuldeep Purohit (kuldeeppurohit3@gmail.com), Praveen Kandula, Maitreya Suin, A N Rajagopalan

\textbf{Affiliations:} Indian Institute Of Technology Madras, India

\subsection*{Nam team}
\textbf{Title:} U-shaped and Densely Spatial/Channel Residual Attention Networks

\textbf{Members:} Nam Hyung Joon (013107nam@naver.com), Yu Seung Won

\textbf{Affiliations:} Image Communication \& Signal Processing Laboratory, Hanyang University, in Seoul, Korea

\subsection*{CVML team}
\textbf{Title:} Dual Path Residual Attention Network

\textbf{Members:} Guisik Kim (specialre@naver.com), Dokyeong Kwon

\textbf{Affiliations:} CVML, Chung-Ang University

\subsection*{ACVLab-NPUST team}
\textbf{Title:} Densely Shortcuts Connection Network for Real-World Image Super-Resolution

\textbf{Members:} Chih-Chung Hsu (cchsu@mail.npust.edu.tw), Chia-Hsiang Lin (chiahsiang.steven.lin@gmail.com)

\textbf{Affiliations:} Department Management Information Systems, National Pingtung University of Science and Technology \& Department of Electrical Engineering, National Cheng-Kung University.

\subsection*{SeeCout team}
\textbf{Title:} Un-paired Real World Super-resolution with Degradation Consistency

\textbf{Members:} Yuanfei Huang (yf\_huang@stu.xidian.edu.cn), Xiaopeng Sun, Wen Lu, Jie Li, Xinbo Gao

\textbf{Affiliations:} Xidian University

\subsection*{Image Specific NN for RWSR team}
\textbf{Title:} Image Specific Neural Networks for Real World Super Resolution

\textbf{Members:} Sefi Bell-Kligler (sefibk@gmail.com)

\textbf{Affiliations:} The Weizmann Institute of Science, Israel
	
	{\small
		\bibliographystyle{ieee_fullname}
		\bibliography{references}
	}
	
\end{document}